\documentclass[letterpaper, 10 pt, conference]{ieeeconf}
\usepackage{graphicx}
\usepackage{cite}
\usepackage{algorithmic}
\usepackage[ruled,linesnumbered]{algorithm2e}
\usepackage{amsmath}  
\usepackage{amssymb}  
\usepackage{multirow}
\usepackage{url}
\usepackage{authblk}
\usepackage{etoolbox}
\IEEEoverridecommandlockouts                              

\title{\LARGE \bf
SimWorld: A Unified Benchmark for Simulator-Conditioned Scene Generation via World Model
}
\author{
    Xinqing Li\textsuperscript{1,2}, Ruiqi Song\textsuperscript{2,3}, Qingyu Xie\textsuperscript{1}, Ye Wu\textsuperscript{1,2}, Nanxin Zeng\textsuperscript{1,2}, Yunfeng Ai\textsuperscript{1,2,\textdagger}\\
    \thanks{
        This work was supported by the National Key Research and Development Program of China, project 3 under Grant 2022YFB4703703. (X. Li and R. Song contributed equally to this work.)
    }
    \thanks{\textdagger\ \textbf{Corresponding author}:Y. Ai. \texttt{aiyunfeng@ucas.ac.cn}}
    \thanks{
        \textsuperscript{1} The School of Artificial Intelligence, University of Chinese Academy of Sciences, Beijing 100049, China \texttt{\{lixinqing22, xieqingyu22, wuye23, zengnanxin24\}@mails.ucas.ac.cn}
    }
    \thanks{\textsuperscript{2} Waytous Inc., Qingdao 266109, China}
    \thanks{
        \textsuperscript{3} The State Key Laboratory of Multimodal Artificial Intelligence Systems, Institute of Automation, Chinese Academy of Sciences, Beijing 100190, China \texttt{ruiqi.song@ia.ac.cn}
    }
    
}

\begin{document}
\maketitle

\begin{abstract}
With the rapid advancement of autonomous driving technology, a lack of data has become a major obstacle to enhancing perception model accuracy. Researchers are now exploring controllable data generation using world models to diversify datasets. However, previous work has been limited to studying image generation quality on specific public datasets. There is still relatively little research on how to build data generation engines for real-world application scenes to achieve large-scale data generation for challenging scenes. In this paper, a simulator-conditioned scene generation engine based on world model is proposed. By constructing a simulation system consistent with real-world scenes, simulation data and labels, which serve as the conditions for data generation in the world model, for any scenes can be collected. It is a novel data generation pipeline by combining the powerful scene simulation capabilities of the simulation engine with the robust data generation capabilities of the world model. In addition, a benchmark with proportionally constructed virtual and real data, is provided for exploring the capabilities of world models in real-world scenes. Quantitative results show that these generated images significantly improve downstream perception models performance. Finally, we explored the generative performance of the world model in urban autonomous driving scenarios. All the data and code will be available at \url{https://github.com/Li-Zn-H/SimWorld}.
\end{abstract}

\section{INTRODUCTION}
\begin{figure}[htbp]
  \centering
  \includegraphics[width=0.45\textwidth]{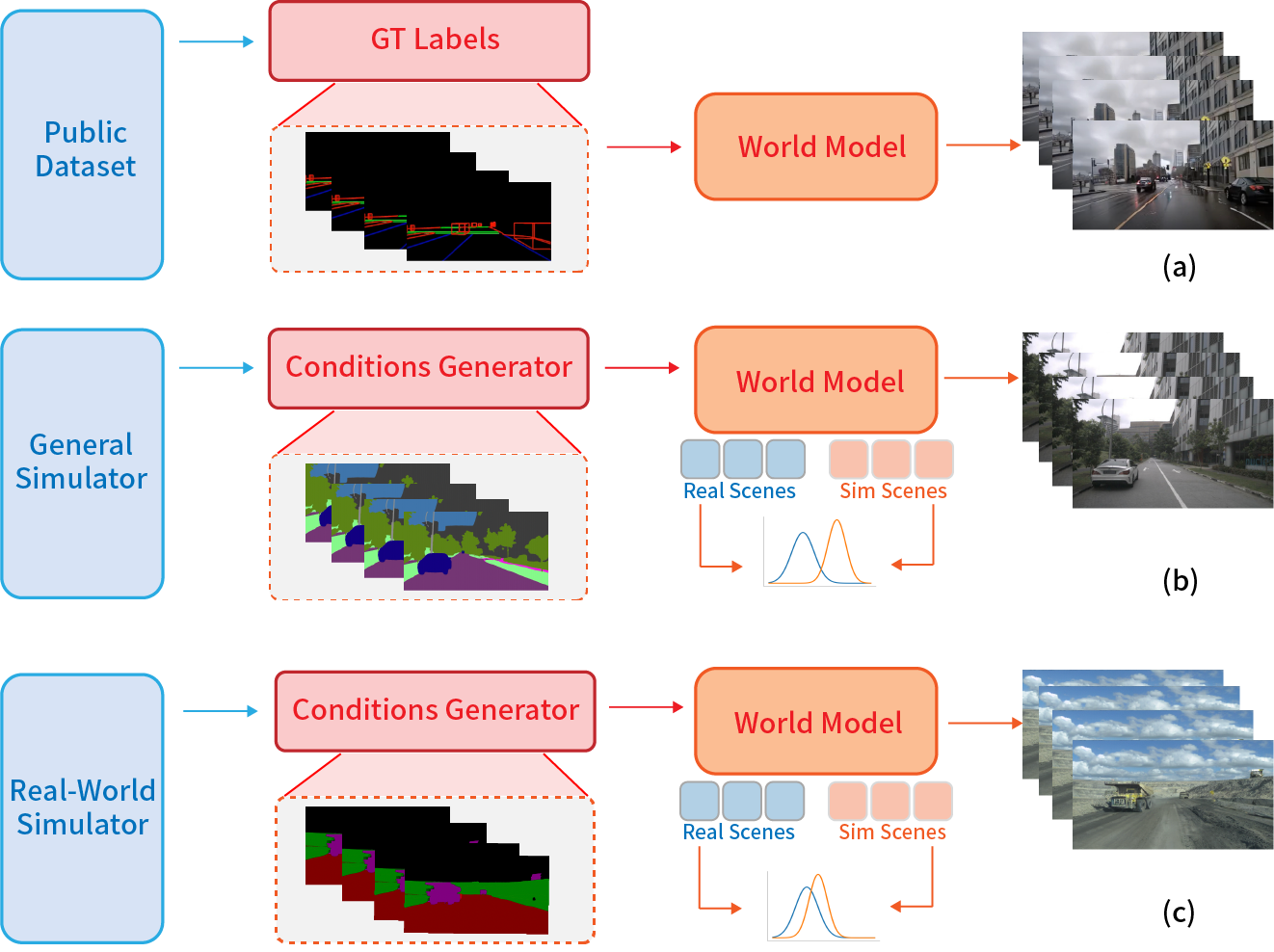}
  \caption{Three data generation paradigms: (a) Conditioned on GT, limiting large-scale data generation; (b) Conditioned on a general simulator, causing significant data distribution differences; (c) Conditioned on a real-world simulator, with minimal distribution differences.}
  \label{fig:fig1}
\end{figure}
In recent years, the rapid progress of autonomous driving has led to the emergence and maturation of methods like end-to-end learning\cite{chen2024end}. These advancements have significantly increased the demand for high-quality datasets, which are essential for driving the technology forward. However, acquiring comprehensive and representative datasets remains a major challenge. A key issue is collecting corner case data\cite{li2022coda}, as rare scenarios—such as extreme weather, sudden road changes, and complex urban settings—are critical for safety but difficult to replicate in real-world conditions. Additionally, the high cost of data annotation poses another obstacle. Fine-grained manual labeling is both time-consuming and expensive, and given the vast data requirements, these factors make data collection a primary bottleneck in autonomous driving development.
To tackle these challenges, researchers are exploring data generation methods to supplement or replace real-world data collection. Some studies utilize simulation environments, such as Virtual KITTI\cite{gaidon2016virtual}, which employs Unreal Engine (UE) to generate driving scenes, and GTA V\cite{richter2016playing}, which collects driving data from its virtual game environment by simulating urban scenarios. Others focus on domain adaptation techniques to enhance the effectiveness of synthetic driving scenes in real-world applications\cite{wang2019weakly, song2023synthetic}.
Although these methods have made progress in improving data generation quality and addressing specific scenarios, existing research still faces certain limitations. Notably, the visual gap between simulation-generated virtual data and real-world data often hinders the generalization performance of autonomous driving systems in real-world applications\cite{hu2023simulation}.
With the emergence of world models, they have increasingly been applied to autonomous driving. Fig. \ref{fig:fig1} presents common application approaches, with the most widely adopted method generating driving scenes conditioned on real data ground truth, as seen in DriveDreamer\cite{wang2023drivedreamer} and DriveDreamer-2\cite{zhao2024drivedreamer}. However, due to the scarcity of corner cases in training data, generated scenes often fail to cover boundary conditions, limiting their ability to create truly novel data. An alternative approach leverages a general simulator as a condition for scene generation, as seen in SimGen\cite{zhou2024simgen}. However, this approach has several limitations. First, scenes generated by general simulators often differ significantly in style from real-world driving scenarios, creating a noticeable distribution gap between generated and real data. Additionally, these simulators lack flexible scene customization, limiting their ability to produce diverse datasets.
\begin{figure*}[htbp]
    \vspace{10pt}
  \centering
  \includegraphics[width=1\textwidth]{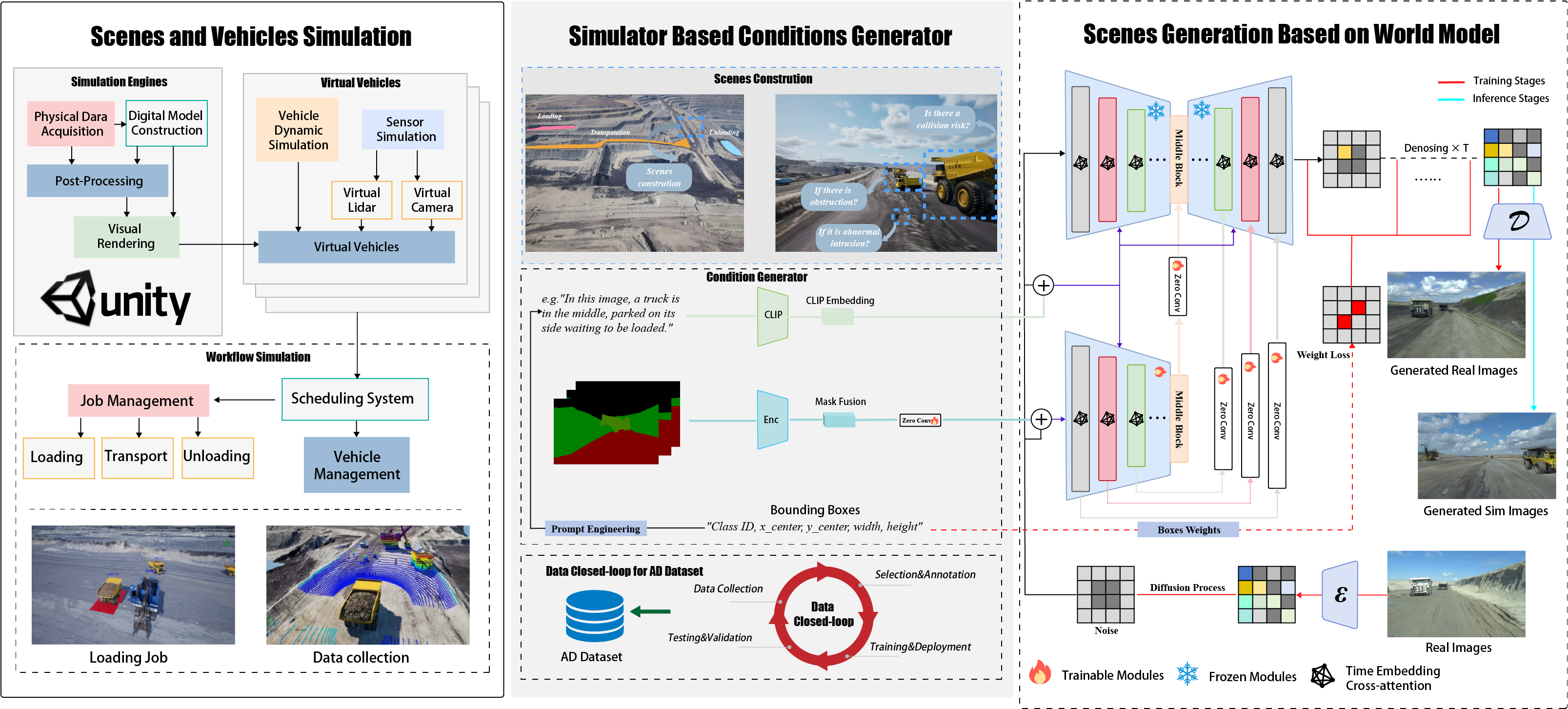}
  \caption{Overall framework of Simworld. Simworld is divided into three main components: Scene and Vehicle Simulation, Simulator Based Conditions generation, and Scenes Generation Based on World Model.}
  \label{fig:framework}
\end{figure*}
As a solution, we propose a benchmark for generating autonomous driving scenarios based on simulator-conditioned labels. Using the parallel mining simulator, we can flexibly create a wide variety of driving scenarios, effectively mitigating the scarcity of data for corner cases and extreme situations. This approach not only achieves greater scene control and label alignment but also substantially reduces the disparity between simulated and real-world data—an improvement difficult to achieve with current methods.

The main contributions of this paper are summarized as follows:
\begin{enumerate}
\item A novel simulator-conditioned scene generation pipeline that integrates the scene simulation power of a simulation engine with the robust data generation of a world model, is proposed.
\item We introduce the first unified benchmark for simulator-conditioned scene generation under real-world conditions. All data and code are open-sourced.
\item Quantitative experiments confirm the quality and diversity of the generated images, showcasing their enhancement of downstream perception tasks. Additionally, Cityscapes is used to further validate our method's effectiveness.
\end{enumerate}

\section{RELATED WORK}
\subsection{Datasets Generated by Virtual Engines}
Since 2012, synthetic datasets like MPI-Sintel\cite{butler2012naturalistic}, derived from the animated film Sintel, have been widely used for optical flow estimation task. With the rise of autonomous driving, interest in synthetic datasets for this field has grown. Virtual KITTI\cite{gaidon2016virtual}, built with the Unity engine, provides multi-task annotated video sequences, while SYNTHIA\cite{ros2016synthia} simulates road environments. Synscapes\cite{wrenninge2018synscapes} is designed to replicate Cityscapes\cite{cordts2016cityscapes}, and synthetic data has also been sourced from virtual games, such as the GTA5 dataset\cite{richter2016playing}, which includes pixel-level semantic annotations from Grand Theft Auto V. Despite advantages like annotation accuracy, condition flexibility (e.g., weather and lighting), and customization for long-tail and edge cases, synthetic datasets still differ significantly in appearance and content from real-world data\cite{song2023synthetic}.

\subsection{Datasets Generated by Deep Generative Models}
The Variational Autoencoder (VAE)\cite{kingma2013auto} is one of the earliest and most widely used generative models. It maps input data to a probabilistic latent space, sampling from it to generate diverse outputs. However, VAEs often produce blurry, low-quality images, making them unsuitable for high-resolution, complex driving scene datasets.

The advent of Generative Adversarial Networks (GANs) revolutionized image generation by leveraging adversarial learning to produce images , offering significant improvements in clarity and detail over VAEs. This breakthrough led to extensive research on GAN-based autonomous driving scene generation. Models like GauGAN\cite{tang2020dual}, SelectionGAN\cite{tang2019multi}, and MaskGAN\cite{lee2020maskgan} focus on creating realistic driving environments. However, despite their ability to generate detailed images, GANs suffer from training instability and poor generalization, making them challenging to train and limiting adaptability to diverse scenarios.

Diffusion models represent a major breakthrough in generative modeling, providing an alternative to traditional methods. Unlike adversarial training, they gradually add noise to data and then restore it using a denoising network. The Denoising Diffusion Probabilistic Model (DDPM)\cite{NIPS20145ca3e9b1}, for instance, generates high-resolution images with stable training, overcoming GANs' instability issues. However, their high computational cost and slow processing speed limit their efficiency for large-scale data generation.

Latent Diffusion Models (LDMs)\cite{rombach2022high} improve efficiency by operating in a compressed latent space, significantly reducing computational costs. Various LDM variants have been explored for generating driving scenarios. For instance, Wang et al. introduced DriveDreamer, training the model to understand road structures and infer future scenes based on driving actions\cite{wang2023drivedreamer}. Zhao et al. further developed DriveDreamer-2, fine-tuning a large language model (LLM) to generate BEV trajectories from user-provided text prompts, enabling personalized driving scene videos\cite{zhao2024drivedreamer}.
These studies highlight the great potential of diffusion models in autonomous driving data generation but also reveal key limitations. First, the training datasets lack extreme cases, leading to generated scenes that do not adequately cover corner cases. Second, most research prioritizes data quality over scalability, limiting large-scale data generation.
Additionally, we note that concurrent research has been exploring simulator-based scene generation. They introduced SimGen\cite{zhou2024simgen}, which leverages MetaDrive\cite{li2021metadrive} and ScenarioNet\cite{li2023scenarionet} as simulation platforms. While both simulators allow the addition of environmental vehicles according to specific requirements, this process involves writing corresponding attributes and code, which limits the flexibility of environment customization. As a result, these simulators mainly replicate existing dataset scenes, limiting their adaptability. These constraints hinder large-scale, diverse dataset generation. Thus, developing 1:1 real-world simulators and generating large-scale data from customized simulated environments remains an emerging and underexplored research area.

\section{FRAMEWORK}
SimWorld is a pipeline designed to generate realistic surface mining scenes from the virtual engine PMWorld\cite{10316640}. Its architecture consists of two main components: training and inference. As shown in Figure \ref{fig:framework}, SimWorld’s training step leverages multimodal features from real mining data, including detection boxes, natural language descriptions, segmentation masks, and pixel dimensions, as control conditions for the model. These inputs guide tensor generation in latent space and ensure that the generated scenes remain accurate and consistent with real-world mining environments.
\subsection{Scenes and Vehicles Simulation}
\label{Simulator}
In our previous work, we proposed PMWorld, a mining autonomous driving parallel testing platform. To ensure consistency and visual realism between the virtual mine and the physical site, the construction process follows these key steps:
\begin{enumerate}
\item \textbf{Scenario Engineering}: We collected physical data via field surveys and drones, built digital models in a virtual engine, and achieved high-fidelity representation of the surface mining area through post-processing and visual rendering.
\item \textbf{Vehicle Modeling}: We constructed dynamic models of various vehicles using vehicle dynamics models, accurately recreated the visual models of the vehicles at a $1:1$ scale through 3D modeling software, and validated the consistency between the digital models and the physical vehicles through rigorous testing standards.
\item \textbf{Sensor Simulation}: The virtual mine's vehicle sensors, including LiDAR, cameras, inertial navigation, and GPS, replicate real-world sensors and transmit data to the vehicle's dynamic module and dispatch system via CAN bus or Ethernet.
\item \textbf{Hardware Components}: The hardware system consists of three main components: the truck domain controller, the excavator-truck collaborative controller, and the server cluster. The truck domain controller is the primary onboard computing unit, while the excavator-truck controller coordinates excavator-truck operations. The server cluster, a cloud-based high-performance computing center, acts as the central hub for fleet management and safety dispatch
\end{enumerate}
With these components, we present PMWorld, a parallel testing platform for autonomous mining. It simulates mining environments and generates virtual mining data, rapidly providing large-scale, accurately labeled data for generative models.

\subsection{Simulator Based Conditions Generator}
\label{Simulator to Condition}
With the Simulator proposed in Section \ref{Simulator}, we extract condition information, which serves as model inputs to guide the generation process and ensure the generated scenes align with preset conditions. Before this, we collected the PMScenes dataset\cite{ai2024pmscenes} using cameras, LiDAR, and other sensors mounted on virtual vehicles. This dataset includes labels for semantic segmentation, depth estimation, object detection, and 3D point clouds. Various operational environments and working conditions were simulated, including intersections, slopes, parking, following, overtaking, and loading operations. To test the autonomous driving system’s adaptability, both dynamic and static obstacles were introduced. To further capture the complexities of autonomous driving, we simulated extreme weather conditions such as rain, blizzards, fog, and dust storms. Data was collected using virtual cameras mounted on virtual mining trucks at a frequency of 2Hz and a resolution of 1920 $\times$ 1200. All PMWorld data includes timestamp-aligned labels, which can support downstream perception tasks or subsequent generation tasks..

\subsection{Scenes Generation Based on World Model}
\label{generation}
\textbf{Learning within the condition information:} To generate real-world scenes from control conditions, we follow the training process outlined on the right side of Fig. \ref{fig:framework}. Specifically, we utilize two versions of ControlNet\cite{zhang2023adding} at different scales, based on Stable Diffusion\cite{rombach2022high} and Stable Diffusion XL \cite{podell2023sdxl}. The model is implemented using a denoising U-Net\cite{ronneberger2015u} architecture.

Let $x_0 \in X$ represent the latent features of the data distribution $p(x)$. During training, progressive noise is added to $x_t$ over time $t \in [0,1]$, gradually converting $x_t$ into Gaussian noise. This process follows a forward Stochastic Differential Equation (SDE)\cite{ho2020denoising}:
\begin{equation}
    \label{eq:forward}
    x_t = \alpha_t x_0 + \beta_t \epsilon, \epsilon \sim \mathcal{N}(0,\mathbf{I}),x_0 \sim p(x),
\end{equation}
where $x_t$ is the data state at timestep $t$, $\alpha_t$ and $\beta_t$ are time-dependent scaling factors.

The denoising process reverses diffusion, estimating the noise $\epsilon_{\theta}$ at each timestep with a neural network and progressively removing it to produce a clear image:
\begin{equation}
    \label{eq:denosing}
    x_{t-1} = \frac{1}{\sqrt{\alpha_t}}(x_t-\frac{\beta_t}{\sqrt{1-\overline{\alpha_t}}}\epsilon_{\theta}(x_t,t,c))+\sigma_t\epsilon,
\end{equation}

where $\epsilon_{\theta}$ is the denoising network (U-Net), $c$ represents condition information, and $\sigma_t$ is a parametric factor that ensures diversity.

In SimWorld, we employ the ControlNet architecture, which guides the denoising process by adding additional control signals:
\begin{equation}
    \label{eq:controlnet}
    y_c = \mathcal{F}(x;\Theta)+\mathcal{Z}(\mathcal{F}(x+\mathcal{Z}(c;\Theta_{Z1});\Theta_c);\Theta_{Z2}),
\end{equation}
where $x$ and $y_c$ represent the input and output feature maps, $\mathcal{F}$ is the neural network block, $\mathcal{Z}$ denotes the $1 \times 1$ zero convolutional layer $\Theta_c$ and $\Theta_{Zi})$ are the trainable parameters of the ControlNet and zero convolution layer.

To capture richer vehicle details and prioritize foreground vehicles, we introduce DynamicForegroundWeightLoss. This approach utilizes a cosine scheduler to gradually adjust loss weights, enhancing both training stability and effectiveness. The procedure is described in Algorithm \ref{algo1}.
\begin{algorithm}
    \caption{Pseudo-code for Dynamic Foreground Weight}
    \label{algo1}
    \begin{algorithmic}[1]
        \STATE \textbf{Input:} Image $x$, bounding box $j$ at step $t$: $\mathbf{b}_{t}^{j}$, current step: $t$, total steps: $T$, min weight: $w_{min}$, max weight: $w_{max}$, current weight: $w_t$, training threshold: $\eta$
        \STATE \textbf{Output:} Weight matrix: $w(\mathbf{b}_{t})$
        \STATE Initialize $w(\mathbf{b}_{t}) \leftarrow$ ones matrix of shape $(bs, 1, h, w)$
        \FOR{$t = 1$ \textbf{to} $T$}
            \IF{$\frac{t}{T} \leq \eta$}
                \STATE $w_t^j \leftarrow w_{min} +\frac{1-\cos{(\frac{t/T}{\eta}\pi)}}{2}(w_{max}-w_{min})$
            \ELSE
                \STATE $w_t^j \leftarrow w_{max} -\frac{1-\cos{(\frac{t/T-\eta}{1-\eta}\pi)}}{2}(w_{max}-w_{min})$
            \ENDIF
            \FOR{each bounding box $j$ in $\mathbf{b}_t$}
                \STATE Extract coordinates $x_1, y_1, x_2, y_2$ from $j$
                \STATE Set the region $[x_1:x_2, y_1:y_2]$ of the weight matrix $w(\mathbf{b}_{t})$ to $w_t^j$
            \ENDFOR
        \ENDFOR
        \RETURN $w(\mathbf{b}_t)$
    \end{algorithmic}
\end{algorithm}

Algorithm \ref{algo1} schedules the foreground weight in two phases: a fast increase followed by a gradual decrease. Initially, the weight rises quickly to focus the model on key targets and speed up optimization. Once it peaks, the model captures key foreground features. The subsequent reduction of the weight prevents over-reliance on the foreground, promoting a balance between foreground and background while refining image details. This approach improves image quality and ensures a balanced generation process. Since SimWorld operates in latent space, we map the weight matrix into this space using bilinear interpolation to maintain smoothness and consistency. This process generates the foreground optimization matrix $w(\mathbf{b}_t)$ while optimizing the model through weighted diffusion loss:
\begin{equation}
    \label{eq:loss}
    \forall t,\min _{\theta} \mathbb{E}_{t,x_t,c,\epsilon}\left[w\left(\mathbf{b}_{t}\right) \cdot\left\|\epsilon-\epsilon_{\theta}\left(\mathbf{x}_{t} ; \mathbf{c}, t\right)\right\|_{2}^{2}\right],
\end{equation}
where $(\mathbf{x}_{t} ; \mathbf{c}, t)$ denotes the noise predicted by the model.

\textbf{Condition information processing:} To enhance the use of bounding box information, we developed a prompt engineering method (see Fig. \ref{fig:framework}), which converts detection labels into natural language descriptions of the mining scene. This approach leverages textual information, offering better integration with the diffusion text-to-image model than simple bounding boxes.

We use a text encoder to process the transformed text, concatenating its outputs along the channel dimension to capture layered semantic information. This enhances the model's expressiveness and image diversity. Additionally, the encoder processes segmentation maps and pixel data, ensuring the model effectively incorporates structural information for more accurate, condition-consistent outputs.

\textbf{Training and freezing modules:} We replicating and freezing the original U-Net parameters $\theta$, from the diffusion model. Training is performed on the copied parameters, with input passed through trainable, zero-initialized convolutional layers. This preserves the model’s generative capabilities while enhancing performance, yielding higher generation quality and finer detail.

\textbf{Simulation-to-reality conversion:} The inference model mirrors the training model, with simulator-collected conditions replacing the real-world ones used during training. These conditions are sampled from random noise and gradually transformed into scene images. We employed the Denoising Diffusion Implicit Models (DDIM)\cite{song2020denoising}, a faster scheduling method from diffusion models, to accelerate sampling. DDIM uses a non-Markovian process, reducing the number of sampling steps while preserving high generation quality. The denoising sampling formula is as follows:
\begin{equation}
    \label{eq:DDIM}
    \scalebox{0.85}{$\displaystyle  x_{t-1}=\sqrt{\overline{\alpha}_{t-1}}\left(\frac{x_t-\sqrt{1-\overline{\alpha_t}}\epsilon_\theta(x_t; \mathbf{c},t)}{\sqrt{\overline{\alpha_t}}}\right)+\sqrt{1-\alpha_{t-1}}\cdot\eta,$}
\end{equation}
Where $\eta$ is a random noise term that controls the noise scale in the sampling process. When $\eta=0$, the sampling becomes deterministic, producing high-quality images with fewer steps.

\begin{figure*}[!ht]
    \vspace{10pt}
  \centering
  \includegraphics[width=0.95\textwidth]{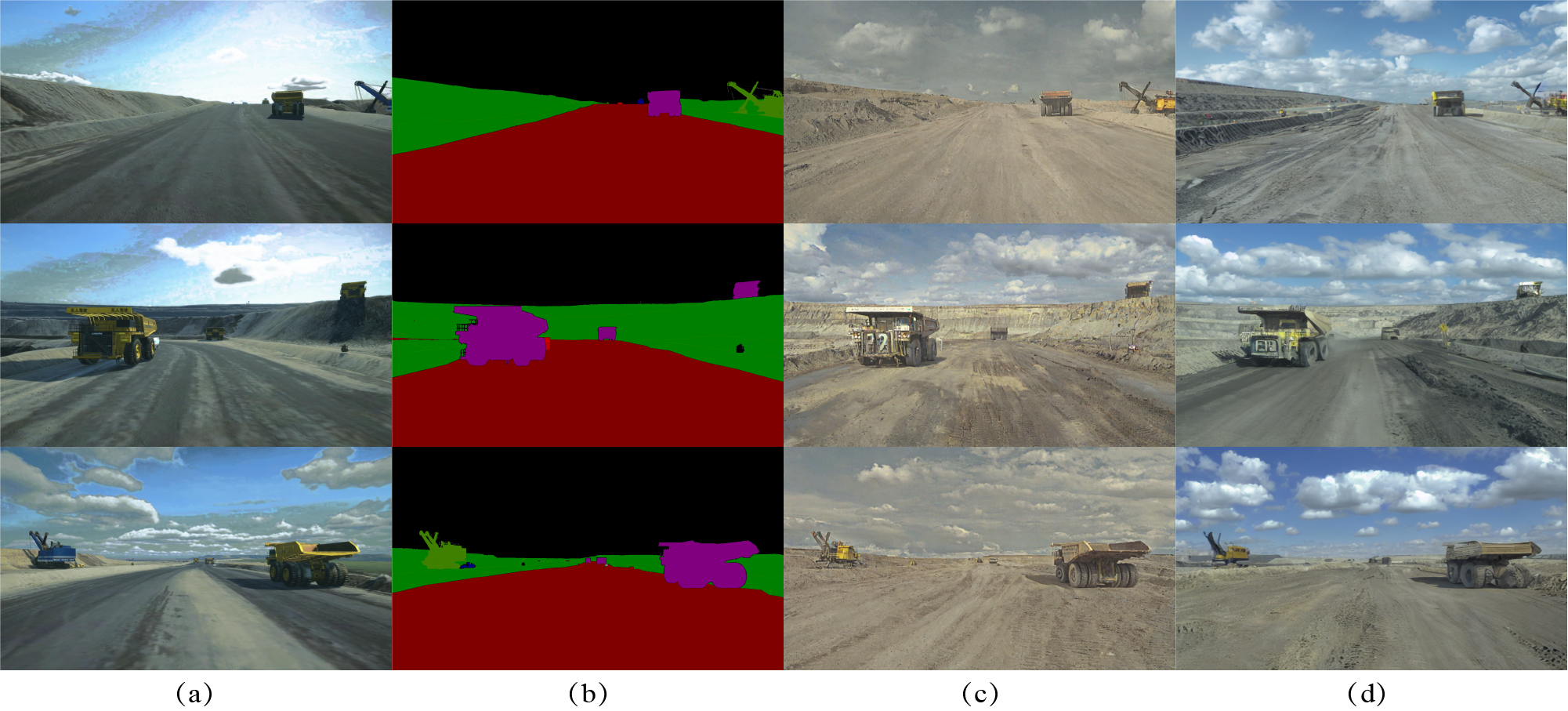}
  \caption{The generated results on PMScenes: (a) represents the simulated image, (b) the segmentation mask, (c) the generated scene by SimWorld XL, and (d) the generated scene by SimWorld.}
  \label{fig:sample}
\end{figure*}

\subsection{Experiment Setting}
\label{experiment setting}
\textbf{Dataset:} The training data is derived from the real mining dataset AutoMine\cite{li2022automine} and augmented to 32k samples through techniques like horizontal flipping and hue adjustments. Each sample includes a scene image, segmentation mask, detection boxes, and prompt-generated descriptions. Inference data, drawn from the PMScenes dataset (11k samples), is preprocessed to match the real mining data format.

\textbf{Trainging:} To assess the impact of model size on generation quality, we trained two versions: SimWorld and SimWorld XL, with SimWorld XL having three times the parameters of SimWorld. SimWorld was trained for 100 epochs on four 4090 GPUs over 33 hours, using a batch size of 2 per step and an effective batch size of 64 through gradient accumulation. SimWorld XL, trained for 100 epochs on two A100 GPUs over 542 hours, also had a batch size of 2 per step, achieving an effective batch size of 32. Both models employed Exponential Moving Average (EMA), OneCycle learning rate scheduling, and AdamW optimization, with the learning rate ranging from 2e-5 to 2e-4.

\textbf{Evaluation:} We evaluated image quality using Frechet Inception Distance (FID)\cite{heusel2017gans} and assessed image diversity with $D_{pix}$. FID calculates the distance between feature vectors of generated and real images by using a pre-trained InceptionV3\cite{szegedy2016rethinking} model to extract features and then computing the Fréchet distance. For diversity, we measured the standard deviation of pixel values, where values closer to those of real images suggest that the style and color of the generated images are more similar to real-world scenes. We calculated the FID score to compare real data with generated images from simulated data, as shown in Tab. \ref{table:FID}.

Additionally, We used the PMScenes methodology to assess the impact of synthetic data on perception models, evaluating performance with standard metrics for detection and segmentation tasks: mean Average Precision (mAP) and mean Intersection over Union (mIoU). mAP evaluates object detection models by calculating precision and recall at various IoU thresholds, then integrating the area under the precision-recall curve to obtain the Average Precision (AP) for each class. IoU is used to measure the performance of image segmentation models, reflecting the overlap between the model's predicted results and the ground truth annotations.

\section{EXPERIMENTS}
\begin{figure*}[!ht]
\vspace{10pt}
  \centering
  \includegraphics[width=0.95\textwidth]{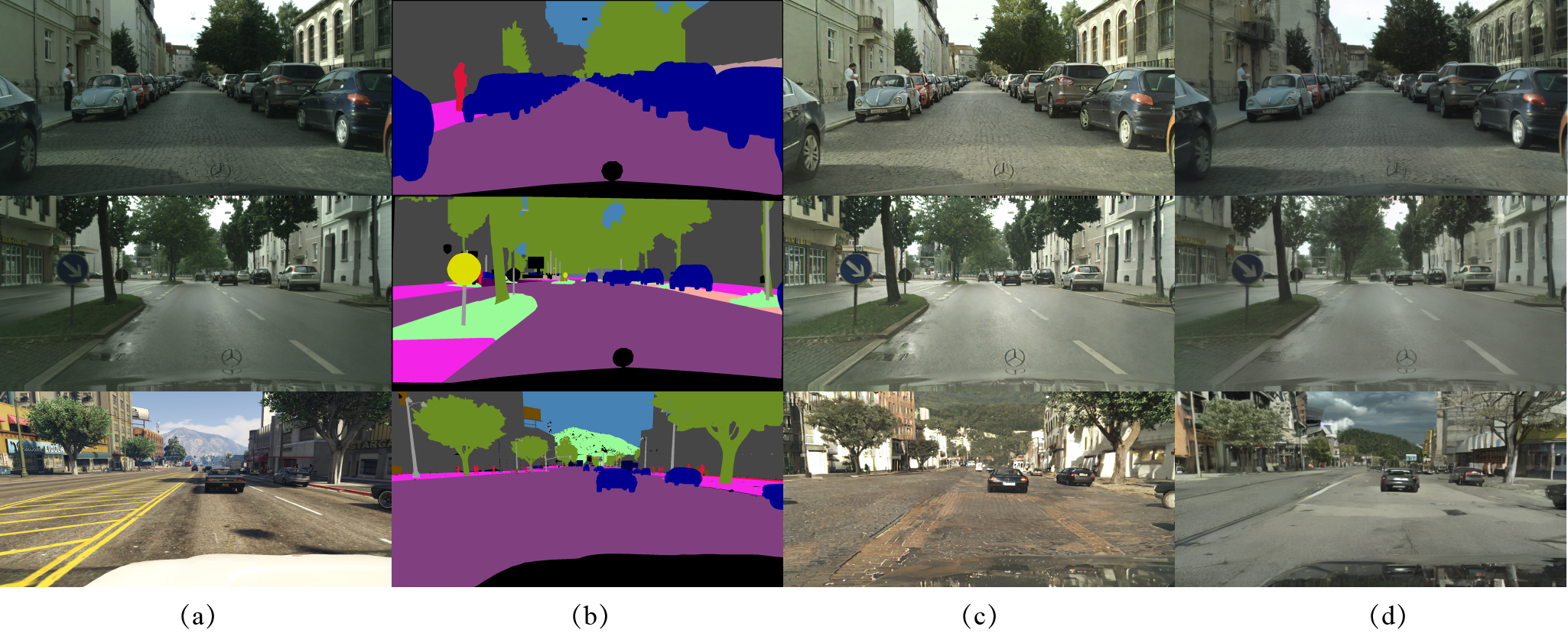}
  \caption{The generated results for urban street: The first two rows show Cityscapes data, while the last row shows synthetic data from GTA5. (a) represents the real-world scene, (b) the segmentation mask, (c) the driving scene generated by SimWorld XL, and (d) the driving scene generated by SimWorld.}
  \label{fig:sample2}
\end{figure*}

\begin{table}[ht]
\centering
\caption{COMPARISONS OF GENERATE QUALITY AND VARIETY}
\label{table:FID}
\renewcommand{\arraystretch}{1.2} 
\setlength{\tabcolsep}{11pt}{
\begin{tabular}{cccc}
\hline\hline
Benckmark                  & Method      & FID$\downarrow$   & $D_{pix}$ \\ \hline
\multirow{5}{*}{AutoMine}  & AutoMine\cite{li2022automine}    & -   & 43.83  \\
                           & PMScenes\cite{ai2024pmscenes}    & 73.45 & 25.79  \\
                           & ProCST\cite{ettedgui2022procst}      & 62.11 & 11.24  \\
                           & SimWorld    & \textbf{33.96} & \textbf{29.08}  \\
                           & SimWorld XL & 36.11 & 26.87  \\ \hline
\multirow{5}{*}{Cityscapes} & Cityscapes\cite{cordts2016cityscapes}   & -    & 37.78  \\
                           & GTA 5\cite{richter2016playing}       & 89.32 & 56.76  \\
                           & ProCST\cite{ettedgui2022procst}      & 81.68 & 67.64  \\
                           & SimWorld    & \textbf{51.93} & \textbf{37.83}  \\
                           & SimWorld XL & 52.83 & 38.84  \\ \hline\hline
\end{tabular}}
\end{table}

This section reviews the model's generative performance and the impact of synthetic data on perception tasks. As shown in Fig. \ref{fig:sample}, SimWorld XL generates more detailed images, while both models at different scales align well with the provided labels. Beyond mining scenes, we also explored urban data. The Cityscapes dataset, with only 3.4k training images, is inadequate for a model of this scale, yet the results in Fig. \ref{fig:sample2} highlight the potential of our approach.

\subsection{Quality and Diversity}

\begin{table*}[!ht]
\vspace{10pt}
\caption{EVALUATION RESULTS OF DIFFERENT ALGORITHMS on 2D OBJECT DETECTION BENCHMARS}
\label{table:objection}
\renewcommand{\arraystretch}{1.2}
\centering
\setlength{\tabcolsep}{7pt}
\begin{tabular}{c c c c c c c c c c c c c}
\hline
\hline
  & & &\multicolumn{10}{c}{METRICS}\\
MODEL&BACKBONE&INPUT& \multicolumn{5}{c}{mAP50(\%)}& \multicolumn{5}{c}{mAP(\%)}\\
& & &RI&PTP&PTS&PTG&MPS&RI&PTP&PTS&PTG&MPS\\
\hline
Faster R-CNN\cite{Ren_2017}&ResNet-r50-FPN&640 * 640
& 45.9 & 42.7 & 48.9& \textbf{50.2} & 47.1 & 70.2 & 68.0 & 71.0 & \textbf{72.3} & 70.6 \\
SSD\cite{Liu_2016}&VGG-16&640 * 640 
& 56.5 & 60.1 & 61.2 & \textbf{63.4} & 60.9 &79.7&82.5&83.6& \textbf{85.1} &84.0\\
YOLOv5\cite{yolov5}&Darknet-53&640 * 640
& 44.8 & 52.4 & 58.4& \textbf{59.7} & 56.3  & 70.9 & 78.2 & 81.6& \textbf{82.0} & 79.3 \\
DETR\cite{zhu2020deformable}&ResNet-50&640 * 640 
&44.6 & 51.7 & 53.8& \textbf{55.2} & 48.6 &72.3&79.3&80.1& \textbf{81.9} &79.5\\
DiffusionDet\cite{chen2022diffusiondet}&ResNet-50&640 * 640 
& 62.5 & 60.1 & 65.7& \textbf{68.1} & 63.2 &82.4&82.7&84.5& \textbf{86.4} &83.5\\
\hline\hline
\end{tabular}
\end{table*}

We assessed generation quality for mining and urban scenes using FID and $D_{pix}$ metrics, with results shown in Tab. \ref{table:FID}. Evaluating mining (AutoMine) and urban (Cityscapes) scenes, we used PMScenes and GTA 5 as inputs for all models. As the table shows, our method achieves a data distribution closer to real-world scenes and exhibits greater pixel-level diversity, aligning more with real environments. This highlights that our approach captures real-world data characteristics better than traditional synthetic images. In comparing models of different scales, we found that while larger models generate more detailed images, their style and diversity slightly lag behind smaller models. We hypothesize that larger models need more data and computational power, with training complexity increasing exponentially. However, this also underscores their potential for better results.

\subsection{Comparative Experiment}
\label{Comparative Experiment}

\begin{table*}[!ht]
\vspace{10pt}
\caption{EVALUATION RESULTS OF DIFFERENT ALGORITHMS on SEMANTIC SEGMENTATION}
\label{table:segmentation}
\renewcommand{\arraystretch}{1.2}
\centering
\setlength{\tabcolsep}{5pt} 
\begin{tabular}{c c c c c c c c c c c c c c c c c}
\hline
\hline
 & &\multicolumn{12}{c}{mIOU(\%)}\\
MODEL & INPUT & \multicolumn{5}{c}{Foreground}& \multicolumn{5}{c}{Background} & \multicolumn{5}{c}{ALL}\\
 & & RI & PTP & PTS & PTG & MPS & RI & PTP & PTS & PTG & MPS & RI & PTP & PTS & PTG &MPS \\
\hline
OCRNet\cite{YuanCW20} & 512*512  & 31.6 & 31.8 & 35.5 & \textbf{50.8}& 50.6 & 81.9 & 80.6 & 83.8& \textbf{92.5} &  87.0      &54.0 &53.5 & 57.0& \textbf{69.7} & 66.8\\
PSPNet\cite{zhao2017pspnet}&512*512&28.2&35.4&36.7& \textbf{54.3}&53.9  & 84.1  & 83.7 & 83.6 & \textbf{93.4}& 87.9     & 53.0 & 56.9 & 57.5& \textbf{71.7} & 69.0\\
DeepLabV3\cite{chen2017rethinking}&512*512 &37.5&42.3&44.4& \textbf{56.7} & 56.0   & 84.5 & 83.8 & 84.5& \textbf{95.8} & 88.3 &58.4 &60.7 &62.2 & \textbf{74.1}&70.4\\
BiSeNetV2\cite{yu2021bisenet}&1024*1024 &24.3&32.7&34.1&\textbf{53.5}&50.2   & 81.0 & 83.0 & 83.1& \textbf{88.9} & 81.5 &49.5 &55.1 &55.9& \textbf{69.2} &64.1\\
U-Net\cite{ronneberger2015u}&512*512&27.1&36.1&36.4& \textbf{54.8} &54.1    & 84.0 & 84.2 & 84.1& \textbf{91.0} & 87.8 &52.4 &57.5 &57.6& \textbf{70.9}&69.1\\

\hline\hline
\end{tabular}
\end{table*}
As the images generated by SimWorld better align with real-world data distributions, we used them for comparative downstream task experiments. To assess the impact of synthetic data on perception models, we adopted the experimental setup from PMScenes. The strategy is as follows:

\textbf{RI (Random Initialization)}: Perception model parameters and biases are randomly initialized and trained directly on AutoMine.

\textbf{PTP (Pre-trained with Public Data)}: Perception model parameters and biases are pre-trained on the KITTI dataset \cite{kitti} and fine-tuned on AutoMine.

\textbf{PTS (Pre-trained with Synthetic Data)}: Perception model parameters and biases are pre-trained on the synthetic dataset PMScenes and fine-tuned on AutoMine.

\textbf{MPS (Mixed Training)}: Perception models are trained using a mix of synthetic data (PMScenes) and real data (AutoMine).

\textbf{PTG (Pre-trained with Generated Images)}: Perception model parameters and biases are pre-trained on generated images and fine-tuned on AutoMine.

\subsubsection{Detection}
To assess the quality of generated images in detection tasks, we validated multiple object detection algorithms, including YOLOv5 \cite{yolov5}, SSD \cite{Liu_2016}, DiffusionDet \cite{chen2022diffusiondet}, DETR \cite{zhu2020deformable}, and Faster R-CNN \cite{Ren_2017}, ensuring the results' applicability and reliability. All experiments followed the methods outlined in Section \ref{Comparative Experiment}. For YOLOv5 and SSD, we used the SGD optimizer with learning rates of $1.0 \times 10^{-2}$ and $2.0 \times 10^{-3}$, respectively. Faster R-CNN used the Adam optimizer with a learning rate of $3.0 \times 10^{-4}$. DiffusionDet and DETR employed the AdamW optimizer, with learning rates of $1.0 \times 10^{-4}$ and $2.5 \times 10^{-5}$, respectively. The corresponding results are shown in Tab. \ref{table:objection}, with the best performance for each algorithm under different training strategies highlighted in bold. The table shows that the random initialization strategy performs significantly worse than the others. This is due to the random assignment of weights and biases, which requires the model to process more data and undergo longer training to learn meaningful features. The model pre-trained on the KITTI dataset performs much better, as it can leverage learned features from autonomous driving data, improving efficiency. Models pre-trained on PMScenes and those using a mixed strategy outperform the KITTI-based model, as PMScenes offers richer, domain-specific information for mining scenes, aiding better adaptation to mining environments. Despite these improvements, the domain gap between simulated and real images still limits further progress. The best performance comes from the model pre-trained with SimWorld-generated images, as it learns key features from real mining scenes and uses segmentation masks to generate images with real-world styles, boosting detection performance.

\subsubsection{Segmentation}
Similar to the detection experiments, we assessed the model's performance using several popular segmentation algorithms, including OCRNet \cite{YuanCW20}, PSPNet \cite{zhao2017pspnet}, DeepLabV3 \cite{chen2017rethinking}, BiSeNetV2 \cite{yu2021bisenet}, and U-Net \cite{ronneberger2015u}, which are well-validated in segmentation tasks.  We applied the five strategies from Section \ref{Comparative Experiment} to evaluate the impact of synthetic images on segmentation models. To analyze the effect of synthetic data, we calculated segmentation performance metrics for both foreground (five vehicle types) and background (scene elements like roads and barriers). All models were optimized with the SGD optimizer, with BiSeNetV2 using an initial learning rate of $5.0 \times 10^{-2}$ and the others set to $1.0 \times 10^{-2}$.  The results in  Tab. \ref{table:segmentation} show a trend similar to the object detection experiments. Additionally, the SimWorld-pretrained strategy surpasses the PMScenes data approach, further highlighting SimWorld's effectiveness in bridging the simulation-to-real-world gap.

\section{CONCLUSION} 
This paper introduces SimWorld, a novel simulator-conditioned scene generation pipeline that combines simulation engines and world models to create realistic, diverse autonomous driving scenarios. It addresses data scarcity for corner cases and bridges the visual gap between synthetic and real-world data. Our unified benchmark is both feasible and promising, offering new opportunities for large-scale data generation in complex, real-world autonomous driving environments.







\bibliographystyle{IEEEtran}  
\bibliography{IEEEabrv,ref}

\end{document}